\documentclass{article}

\PassOptionsToPackage{numbers, compress}{natbib}
\usepackage[main, final]{neurips_2025}

\usepackage[utf8]{inputenc}
\usepackage[T1]{fontenc}
\usepackage{url}
\usepackage{booktabs}
\usepackage{amsfonts}
\usepackage{nicefrac}
\usepackage{microtype}
\usepackage[table]{xcolor}
\usepackage{graphicx}
\usepackage{amsmath}
\usepackage{mathtools}
\usepackage{enumitem}
\usepackage{subcaption}
\usepackage{placeins}
\usepackage[normalem]{ulem}

\usepackage{xcolor}

\usepackage[pagebackref,breaklinks,colorlinks]{hyperref}
\hypersetup{
    colorlinks=true,
    linkcolor=blue,
    filecolor=magenta,
    urlcolor=blue,
    citecolor=blue,
}

\newcommand{\appref}[1]{\hyperref[#1]{Appendix~\ref*{#1}}}

\newcommand{\R}{\mathbb{R}}

\definecolor{fgcpabrow}{RGB}{237,244,255}

\title{A Regularity Condition Under Which a CPAB Transformation is a Diffeomorphism}

\author{%
  \textbf{Amit Aflalo} \normalfont and 
  \textbf{Oren Freifeld}\\
  \normalfont
  The Stein Faculty of Computer and Information Science,
  Ben-Gurion University of the Negev, Israel
}

\begin{document}

\maketitle

\begin{abstract}
Continuous Piecewise-Affine Based (CPAB) transformations, introduced by Freifeld et al.~in~\cite{cpab,cpab2017}, are defined as the flows obtained by integrating continuous piecewise-affine (CPA) velocity fields. These highly expressive, well-behaved transformations have proven useful in a variety of applications, including nonlinear temporal warping and trainable activation functions in 1D, as well as nonrigid spatial deformations in 2D and 3D. While computational costs were not an issue in 1D, they originally limited the practical expressiveness of CPAB transformations in 2D and, much more severely, in 3D. Our recent work (Aflalo et al.~\cite{Aflalo:NeurIPS:2026}), however, removed this bottleneck, enabling efficient optimization of CPAB transformations with much finer tessellations and substantially greater expressiveness, thereby providing a more practical way than traditional methods for representing and constructing diffeomorphisms in 1D, 2D, and 3D.

There is, however, a mathematical subtlety that was overlooked in the original CPAB papers~\cite{cpab,cpab2017}. CPAB transformations are always homeomorphisms and, in fact, bi-Lipschitz, implying that both the transformations and their inverses are differentiable almost everywhere. However, an additional regularity condition is needed to ensure that these transformations are also diffeomorphisms. The purpose of this note is to remedy this omission by identifying a sufficient regularity condition under which a CPAB transformation is guaranteed to be a diffeomorphism. In 1D and 2D, we show that these conditions can be easily enforced and, moreover, hold with probability one under standard sampling assumptions; we expect analogous behavior in 3D.
\end{abstract}

\section{Introduction}
Throughout this note, a diffeomorphism is a differentiable bijection whose inverse is also differentiable. 
Diffeomorphisms are a natural choice in spatial modeling tasks that require differentiable, invertible, and topology-preserving deformations. 
They are widely used in applications such as registration, spatial alignment, and geometric warping, where folding, tearing, or self-intersection would result in geometrically invalid deformations.
The standard way to construct such transformations is via integration of either a stationary velocity field (SVF)~\cite{svf} or a time-dependent velocity field as in Large Deformation Diffeomorphic Metric Mapping (LDDMM)~\cite{lddmm}. However, in practice, such methods often rely on a dense-grid discretization of the spatial domain. As a result, the number of parameters controlling the diffeomorphisms grows directly with the resolution of the domain (e.g., with the number of pixels in 2D). In modern 2D and especially 3D settings, this leads to high memory use, slow optimization, and difficult optimization landscapes.

Continuous Piecewise-Affine Based (CPAB) transformations, introduced by Freifeld et al.~\cite{cpab,cpab2017}, offer a compelling alternative. Instead of a dense-grid discretization, this approach tessellates the domain into cells (e.g., triangles in the 2D case) and constrains the velocity field to be a continuous piecewise-affine (CPA) function. A CPAB transformation is, by definition~\cite{cpab}, obtained by integrating a CPA velocity field. 
CPAB transformations provide a compact low-dimensional parameterization while still capturing rich nonlinear deformations and have been used in diverse deep learning settings, including 1D applications such as time-series alignment and trainable activation functions~\cite{Cyclic,martinez2022closed,shapira2019diffeomorphic,weber2023regularization,DiTAC,digraf, neifar2024leveraging, kryeem2023personalized, timepoint}, parameter-efficient multi-task fine-tuning~\cite{ditask}, and 2D vision and generative modeling tasks~\cite{hauberg2016dreaming,detlefsen2018deep,schwobel2022probabilistic,shacht2021single,skafte2019explicit,wang2024continuous, hartshorne2025arc}.

CPAB transformations are homeomorphisms and even bi-Lipschitz functions (hence
also almost-everywhere differentiable with an almost-everywhere
differentiable inverse). Typically, but technically not always, CPAB transformations are also diffeomorphisms. 
Under a suitable regularity condition, a CPAB transformation is a diffeomorphism.
The purpose of the present paper
is to provide and explain this condition, as well as to show that in 1D and 2D it usually holds in practice. We expect that it also usually holds in 3D.

\paragraph{Relation to the ODE literature.}
The dependence of ODE flows on initial conditions is a classical
topic~\cite{Hartman2002}. Related questions concerning crossings,
tangencies, and differentiability of flows at switching interfaces
arise in the theory of piecewise-smooth dynamical
systems~\cite{diBernardo2008,Burden2016,DieciLopez2011}.
Some of this literature considers discontinuous vector fields,
whereas the CPA velocity fields considered here are continuous
and Lipschitz. Our focus is on sufficient regularity conditions
that guarantee differentiability of CPAB flows, and on whether
these conditions hold generically for commonly used tessellations.

\paragraph{A Remark on Expressiveness and Scalability.}
Until recently, despite their compact parameterization, CPAB transformations remained hard to scale in practice. The bottleneck lay in computing the loss gradient, \(\nabla_\theta\mathcal L\),
where the standard formulation made optimization prohibitively expensive once tessellations become fine or spatial inputs become large. Consequently, traditional CPAB implementations were
limited to coarse tessellations, restricting their practical expressiveness. 
Fortunately, in work concurrent with this paper~\cite{Aflalo:NeurIPS:2026}, we showed how 
to remove this bottleneck. The consequence was drastic improvements (orders of magnitude) in speed and memory consumption.  Such gains enable efficient optimization of fine-tessellation CPAB diffeomorphisms even in 3D, and make CPAB competitive as a modern primitive for large-scale nonrigid spatial modeling.

\section{The 1D Case}
\label{sec:1d}

We begin with the one-dimensional case, where the regularity issue can be understood explicitly.
The examples below are accompanied by plots of the velocity field, its time-one flow map, and the derivative of that map.
In these plots, vertical dashed lines mark cell boundaries, the gray dotted diagonal denotes the identity map, and black dots in the velocity and flow panels mark stationary points.

Let $\Omega\subseteq\R$ be an interval, tessellated into $C$ smaller intervals, which we will refer to as cells. A continuous piecewise-affine (CPA) velocity field $v:\Omega\to\R$ is affine within each cell and continuous across cell boundaries. Thus, within cell $c$,
\begin{align}
v_c(p)=a_c p+b_c,
\end{align}
where $a_c,b_c\in\R$. The corresponding CPAB transformation, denoted by $\Phi$, is obtained by integrating the stationary ordinary differential equation
\begin{align}
\dot{p}^{(t)}=v(p^{(t)}),\qquad
p^{(0)}=p_0,\qquad t\in[0,1],
\label{eq:1d_ode}
\end{align}
and setting $\Phi(p_0)=p^{(1)}$.
This transformation, from $\Omega$ onto $\Phi(\Omega)\subset \mathbb{R}$, is referred to as the time-one flow map associated with $v$.

Since $v$ is continuous and piecewise affine, it is Lipschitz continuous. Consequently, its flow is well-defined, invertible, and bi-Lipschitz\footnote{A bi-Lipschitz function is an invertible function that is Lipschitz continuous and whose inverse is also Lipschitz continuous.}. 
In particular, $\Phi:\Omega\to\Phi(\Omega)$ is a homeomorphism\footnote{A homeomorphism is an invertible continuous function  whose inverse is also continuous.}, and both $\Phi$ and $\Phi^{-1}$ are differentiable almost everywhere. 
We will now address the question whether $\Phi$ and $\Phi^{-1}$ are differentiable everywhere.
In other words, we ask whether $\Phi$ is a diffeomorphism.

\begin{figure}[!htbp]
\centering
\includegraphics[width=\linewidth]{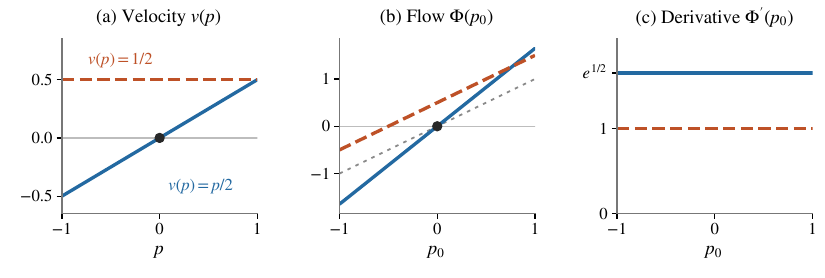}
\caption{\textbf{Affine fields, including the constant case.}
The solid blue curves correspond to $v(p)=p/2$, giving $\Phi(p_0)=e^{1/2}p_0$ and $\Phi'(p_0)=e^{1/2}$.
Although the origin is stationary, the flow is differentiable there.
The dashed orange curves show $v(p)=1/2$ ($a=0$), whose flow is the translation $\Phi(p_0)=p_0+1/2$ with derivative $1$.
Both derivatives are everywhere positive.}
\label{fig:1d_affine}
\end{figure}

\paragraph{An affine velocity field.}
Consider first a single affine velocity field,
\begin{align}
v(p)=ap+b,
\end{align}
where $a,b\in\R$.
The corresponding time-$t$ flow map, $\Phi^{(t)}(p_0)=p^{(t)}$, is given by
\begin{align}
\Phi^{(t)}(p_0)=
\begin{cases}
e^{at}p_0+\dfrac{e^{at}-1}{a}b, & a\neq 0,\\[6pt]
p_0+bt, & a=0.
\end{cases}
\end{align}
In particular, the time-one flow map is
\begin{align}
\Phi(p_0)=\Phi^{(1)}(p_0)=
\begin{cases}
e^a p_0+\dfrac{e^a-1}{a}b, & a\neq 0,\\[6pt]
p_0+b, & a=0.
\end{cases}
\end{align}
and therefore
\begin{align}
\boxed{
\frac{d\Phi(p_0)}{dp_0}=e^a>0.
}
\label{eq:1d_affine_derivative}
\end{align}
Note this result holds regardless of whether the velocity field has a zero in $\Omega$ (i.e., 
a point $x\in\Omega$ where $v(x)=0$). Thus, an affine velocity field always generates a diffeomorphism, and the presence of a zero introduces no regularity issue.
\autoref{fig:1d_affine} illustrates both a field with a zero and a constant nonzero field.

\paragraph{A CPA velocity field without zeros.}
Unlike the affine case, the flow map of a general CPA velocity field has a more complicated expression. Although this expression is available in closed form~\cite{cpab,cpab2017}, we will not need it here. Instead, we derive the derivative of the flow map directly from the defining ODE, which is simpler for studying its regularity.

Consider a CPA velocity field that does not vanish anywhere on $\Omega$. Since $v$ is also  continuous, it has a constant sign on $\Omega$, and every trajectory is a monotonic function (possibly crossing cell boundaries);
that is, if $v$ is positive then the trajectories are monotonic increasing functions (so each point moves rightward)
while if $v$ is negative then the trajectories are monotonic decreasing functions (each point moves leftward).

Recall that the trajectory satisfies
\begin{align}
\dot{p}^{(t)}=v(p^{(t)}),\qquad p^{(0)}=p_0.
\end{align}
Since $v$ is nonzero, we can divide both sides by $v(p^{(t)})$ to obtain
\begin{align}
\frac{\dot{p}^{(t)}}{v(p^{(t)})}=1.
\end{align}

Integrating both sides from $0$ to $t$, using $s$ as the integration variable, gives
\begin{align}
\int_0^t \frac{\dot{p}^{(s)}}{v(p^{(s)})}\,ds
=
\int_0^t 1\,ds
=t.
\end{align}
Now, introducing the new variable $p=p^{(s)}$, so that $dp=\dot{p}^{(s)}\,ds$, we obtain
\begin{align}
\int_{p_0}^{p^{(t)}}\frac{1}{v(p)}\,dp=t.
\end{align}

To establish differentiability of the flow, fix $p_*\in\Omega$ and define
\begin{align}
G(p)=\int_{p_*}^{p}\frac{1}{v(q)}\,dq.
\end{align}
Since $v$ is continuous and nowhere zero, $G$ is continuously differentiable,
with $G'(p)=1/v(p)\neq 0$. Thus, $G$ is strictly monotonic, and the
inverse function theorem guarantees that $G^{-1}$ is continuously differentiable.
The integral identity above can now be written as
\begin{align}
G(p^{(t)})=G(p_0)+t,
\end{align}
or, equivalently,
\begin{align}
\Phi^{(t)}(p_0)=G^{-1}(G(p_0)+t).
\end{align}
Hence, the flow is continuously differentiable with respect to its initial
condition. Differentiating this expression and setting $t=1$ gives
\begin{align}
\boxed{
\frac{d\Phi(p_0)}{dp_0}
=
\frac{v(\Phi(p_0))}{v(p_0)}.
}
\label{eq:1d_derivative_ratio}
\end{align}

\begin{figure}[!htbp]
\centering
\includegraphics[width=\linewidth]{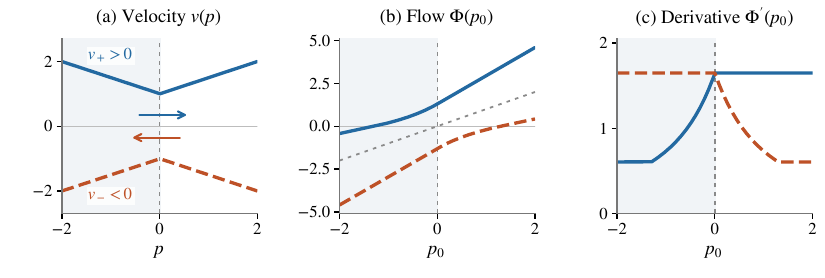}
\caption{\textbf{Nonzero velocity at a cell boundary.}
The fields $v_+(p)=1+|p|/2$ (solid blue) and $v_-(p)=-(1+|p|/2)$ (dashed orange) have unequal slopes across $p=0$, but neither vanishes there.
Arrows indicate the direction of motion; trajectories can cross the interface.
For both fields, the time-one flow is smooth to first order across the interface, and $\Phi'(p_0)=v(\Phi(p_0))/v(p_0)$ is continuous and strictly positive.}
\label{fig:1d_nonzero}
\end{figure}

Since $v$ is continuous and nonzero, the right-hand side is continuous and strictly positive. Importantly, this remains true even when a trajectory crosses one or more cell boundaries, where the derivative of $v$ (as opposed
to the derivative $\frac{d\Phi(p_0)}{dp_0}$) may be discontinuous. 
As a similar analysis holds for the inverse flow (namely, the inverse of $\Phi^{(t)}$), we conclude that $\Phi$, 
the resulting CPAB transformation, is a diffeomorphism\footnote{In fact, even without such analysis of the inverse flow
we can reach the same conclusion: since the derivative of \(\Phi\) is everywhere positive and continuous,
the inverse function theorem establishes that \(\Phi^{-1}\) is continuously differentiable.}.
\autoref{fig:1d_nonzero} shows that a kink in a nonvanishing velocity field does not produce a kink in the flow.

We have just shown that if the CPA field $v$ is nowhere zero, then the derivative of the flow
exists and is continuous. In fact, the same argument applies locally at every $p_0$ provided that $v(p_0)\neq 0$, even if the velocity field has zeros elsewhere. To see this, consider the maximal interval containing $p_0$ on which $v$ does not vanish. By uniqueness of ODE solutions, a trajectory starting at $p_0$ cannot reach a zero of $v$ in finite time. Thus, as long as the flow is defined, the trajectory remains in this interval, where $G$ is well-defined and the preceding argument applies.

\paragraph{Zero boundary conditions with no zeros at interior cell boundaries.}
We now consider the case where the velocity field vanishes at the endpoints of the domain but does not vanish at any interior cell boundary. Zeros strictly inside cells are allowed.

Specifically, let $\Omega=[p_{\min},p_{\max}]$, and suppose
\begin{align}
v(p_{\min})=v(p_{\max})=0.
\end{align}
These boundary points are fixed points of the flow, so the flow preserves the interval.

The presence of these zeros does not prevent the resulting transformation from being a diffeomorphism of the interval. To see this, consider the cell adjacent to $p_{\min}$. Since $v$ is affine in that cell and vanishes at $p_{\min}$, it has the form
\begin{align}
v(p)=a_1(p-p_{\min}).
\end{align}
For initial points sufficiently close to $p_{\min}$, the trajectory remains within this cell for every $t\in[0,1]$, and therefore
\begin{align}
\Phi(p_0)=p_{\min}+e^{a_1}(p_0-p_{\min}).
\end{align}
Hence, the one-sided derivative at the boundary exists and equals
\begin{align}
\Phi'(p_{\min})=e^{a_1}>0.
\end{align}
An analogous argument applies at $p_{\max}$, with the slope of the last cell.

\begin{figure}[!htbp]
\centering
\includegraphics[width=\linewidth]{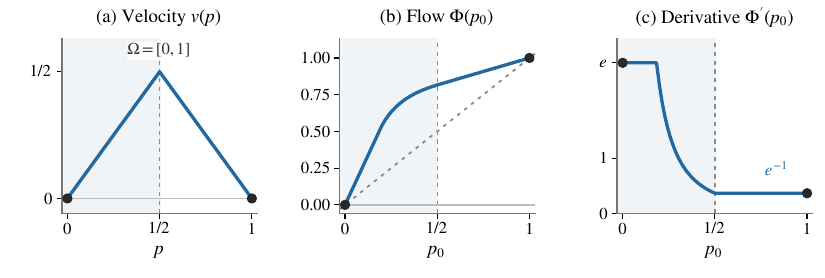}
\caption{\textbf{Zeros at the domain endpoints.}
On $\Omega=[0,1]$, take $v(p)=\min(p,1-p)$, with cells meeting at $p=1/2$.
The endpoints are fixed, while interior points move to the right and remain inside the interval.
The flow has a continuous positive derivative in the interior and positive one-sided endpoint derivatives $\Phi'(0)=e$ and $\Phi'(1)=e^{-1}$.
The nonzero velocity at the interior interface allows smooth passage between cells.}
\label{fig:1d_zero_endpoints}
\end{figure}

Thus, zero boundary conditions introduce no regularity problem. Under the assumptions stated above, the CPAB transformation is a diffeomorphism on the interior of $\Omega$, with positive one-sided derivatives at its endpoints.
An example with two cells is shown in \autoref{fig:1d_zero_endpoints}.

\paragraph{A CPA velocity field with a zero inside a cell.}
Next, suppose that $v$ vanishes at some interior point $p_*$ of a cell $c$, so that $v(p_*)=0$. Since $v$ is affine within this cell, we can write
\begin{align}
v(p)=a_c(p-p_*),\qquad p\in c.
\end{align}
The point $p_*$ is stationary, i.e., $\Phi(p_*)=p_*$. Moreover, for initial points sufficiently close to $p_*$, the trajectories remain within cell $c$ throughout the integration time interval. Consequently,
\begin{align}
\Phi(p_0)=p_*+e^{a_c}(p_0-p_*),
\end{align}
and hence
\begin{align}
\frac{d\Phi(p_0)}{dp_0}=e^{a_c}>0
\end{align}
throughout a neighborhood of $p_*$.

\begin{figure}[!htbp]
\centering
\includegraphics[width=\linewidth]{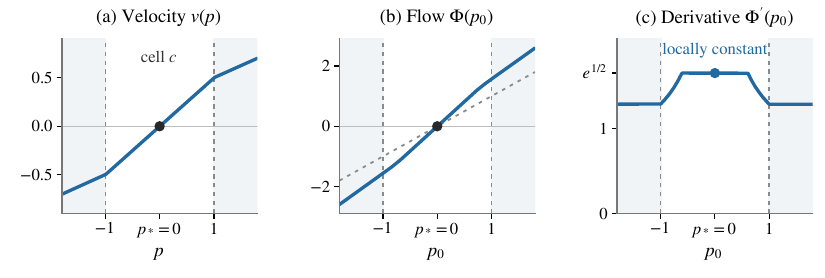}
\caption{\textbf{A zero strictly inside a cell.}
Here $v(p)=p/2$ on the central cell $[-1,1]$, $v(p)=(p-1)/4$ for $p<-1$, and $v(p)=(p+1)/4$ for $p>1$.
The zero $p_*=0$ lies away from the interfaces, where the velocity is nonzero.
For $|p_0|<e^{-1/2}$, trajectories remain in the central cell, so $\Phi(p_0)=e^{1/2}p_0$ and $\Phi'(p_0)=e^{1/2}$.
Thus, the derivative is well-defined at the zero and continuous throughout.}
\label{fig:1d_zero_inside_cell}
\end{figure}

Thus, {although the general derivative formula in~\autoref{eq:1d_derivative_ratio} cannot be applied directly at $p_*$, the flow is differentiable there, with a continuous and strictly positive derivative}. Away from zeros of $v$, that formula remains applicable, including at cell boundaries where the velocity field is nonzero. Therefore, a zero strictly inside a cell introduces no regularity issue, and the resulting CPAB transformation remains a diffeomorphism provided that no zeros occur at problematic cell interfaces.
\autoref{fig:1d_zero_inside_cell} illustrates this local affine behavior around a stationary point.

\begin{figure}[!htbp]
\centering
\includegraphics[width=\linewidth]{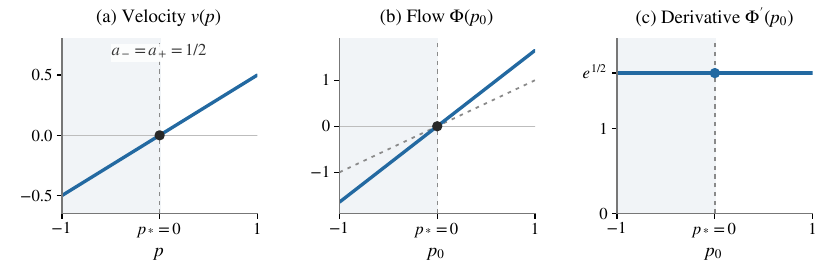}
\caption{\textbf{An interface zero with matching slopes.}
The two cells meet at $p_*=0$, and both affine pieces are $v(p)=p/2$.
The interface therefore introduces no change in the local affine field.
Although $v(p_*)=0$, the flow is $\Phi(p_0)=e^{1/2}p_0$ on both sides, and its one-sided derivatives agree: $\Phi'_-(0)=\Phi'_+(0)=e^{1/2}$.
The stationary point is a differentiable point of the flow.}
\label{fig:1d_matching_slopes}
\end{figure}

A similar argument applies when $v$ vanishes at an interior boundary between two adjacent cells, provided that the slopes of the two affine pieces agree. Indeed, by continuity, the two pieces then define the same affine function in a neighborhood of that boundary. Consequently, the flow is differentiable there, with a continuous and positive derivative given by the exponential of their common slope.
This harmless interface zero is illustrated in \autoref{fig:1d_matching_slopes}.

To summarize the discussion above, as long as $v$ does not vanish at any interior cell boundary where the slopes of the adjacent affine pieces differ, the resulting CPAB transformation is a diffeomorphism. Its derivative is given by
\begin{align}
\boxed{
\frac{d\Phi(p_0)}{dp_0}
=
\begin{cases}
e^a, & \text{if } v(p_0)=0
       \text{ and } v \text{ is affine on a neighborhood of } p_0,\\[6pt]
\dfrac{v(\Phi(p_0))}{v(p_0)}, & \text{if } v(p_0)\neq 0.
\end{cases}
}
\end{align}
where, in the first case, $a$ is the slope of $v$ in a neighborhood of $p_0$, with a one-sided interpretation at the endpoints of $\Omega$.

\paragraph{A zero at a cell boundary.}
We now consider the case responsible for the regularity issue: a zero at an interior cell boundary where the slopes of the adjacent affine pieces differ. Let $p_*$ denote this boundary point, so that
\begin{align}
v(p_*)=0.
\end{align}
By continuity and piecewise affinity, in a neighborhood of $p_*$ we may write
\begin{align}
v(p)=
\begin{cases}
a_-(p-p_*), & p\leq p_*,\\
a_+(p-p_*), & p\geq p_*,
\end{cases}
\label{eq:1d_interface_velocity}
\end{align}
where $a_-$ and $a_+$ are the slopes of the velocity field in the two cells.

The point $p_*$ is stationary, so $\Phi(p_*)=p_*$. For initial points sufficiently close to $p_*$, the trajectories remain within their respective cells throughout the integration interval. Consequently,
\begin{align}
\Phi(p_0)=
\begin{cases}
p_*+e^{a_-}(p_0-p_*), & p_0\leq p_*,\\
p_*+e^{a_+}(p_0-p_*), & p_0\geq p_*.
\end{cases}
\end{align}

The one-sided derivatives at $p_*$ are therefore
\begin{align}
\Phi'_-(p_*)=e^{a_-},
\qquad
\Phi'_+(p_*)=e^{a_+}.
\label{eq:1d_interface_derivatives}
\end{align}

If $a_-\neq a_+$, these derivatives differ, and hence $\Phi$ is not differentiable at $p_*$. Nevertheless, both one-sided derivatives are strictly positive, and $\Phi$ remains a bi-Lipschitz homeomorphism.

\begin{figure}[!htbp]
\centering
\includegraphics[width=\linewidth]{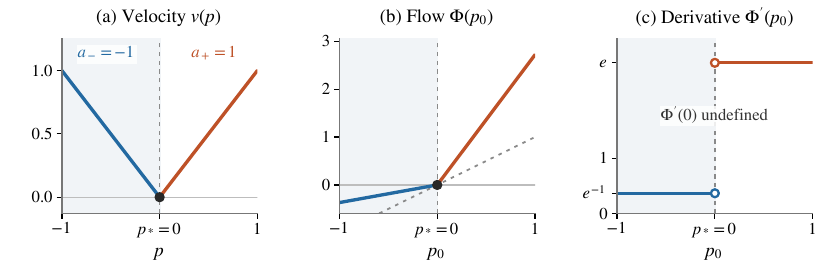}
\caption{\textbf{An interface zero with differing slopes: the failure case.}
For $v(p)=|p|$, the slopes at $p_*=0$ are $a_-=-1$ (blue) and $a_+=1$ (orange).
The time-one map is $e^{-1}p_0$ on the left and $ep_0$ on the right.
It is continuous and strictly increasing, but has a kink at the fixed point.
Open circles mark the unequal derivative limits $e^{-1}$ and $e$; $\Phi'(0)$ does not exist.
The map remains bi-Lipschitz, but is not a diffeomorphism.}
\label{fig:1d_mismatching_slopes}
\end{figure}

This demonstrates that the continuity and Lipschitz continuity of a CPA velocity field do not, by themselves, guarantee that its flow is a diffeomorphism. The difficulty arises when a zero of the velocity field coincides with an interior cell boundary at which the affine slopes differ.
\autoref{fig:1d_mismatching_slopes} makes this failure visible as a kink in the flow and unequal one-sided limits of its derivative.

\paragraph{Conclusion for the 1D case.}
The examples above identify the essential distinction. Zeros of the velocity field are not inherently problematic. 
In particular, zeros in the interior of cells, zeros at interior cell boundaries where the adjacent affine slopes agree, and zeros at the endpoints of a bounded domain are all compatible with diffeomorphic flows. Likewise, discontinuities in the derivative of the velocity field at cell boundaries cause no problem when the velocity is nonzero there.

A sufficient regularity condition in 1D is therefore that, at every interior cell boundary where the affine slopes differ, the velocity field is nonzero. Under this condition, the CPAB transformation is a diffeomorphism (with the usual one-sided interpretation at the endpoints of a closed interval). In fact, the preceding analysis establishes the stronger result that both the transformation and its inverse are continuously differentiable. Thus, the transformation is a $C^1$ diffeomorphism on the interior of the domain.

\textbf{Note that, importantly, this regularity condition is naturally satisfied in practice.} For example, if the free velocities at the tessellation vertices are sampled jointly from an absolutely continuous distribution (e.g., a multivariate Gaussian distribution), then the condition holds almost surely. The same is true if arbitrarily small Gaussian noise is added to the free velocities at the tessellation vertices, while preserving any prescribed boundary conditions. Random initialization followed by gradient-based optimization is also expected to satisfy the condition in typical settings, although this is not guaranteed for arbitrary optimization procedures. In contrast, the two sampling or perturbation procedures described above guarantee that the condition holds with probability one. Thus, for all practical purposes, 1D CPAB transformations are usually diffeomorphisms.   

\FloatBarrier
\section{The 2D Case}
\label{sec:2d}

We now turn to the 2D case. As in 1D, a CPA velocity field is continuous across cell boundaries, but its derivative may be discontinuous there. The question is whether such discontinuities can prevent the resulting CPAB transformation from being a diffeomorphism.

\begin{figure}[!htbp]
\centering
\includegraphics[width=0.85\linewidth]{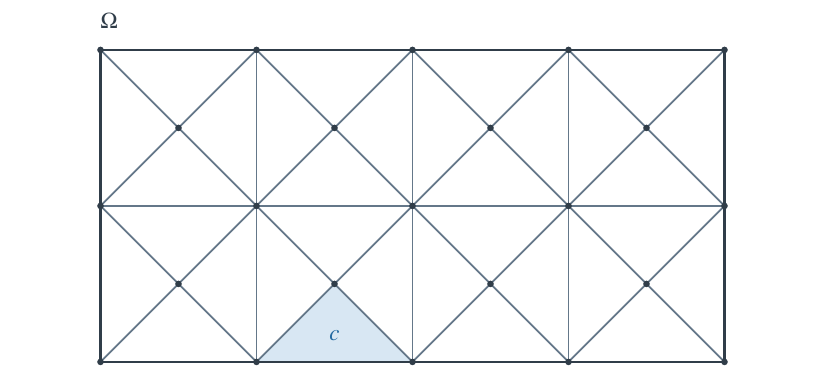}
\caption{\textbf{2D tessellation.}
A $4\times 2$ grid of squares, each divided into four triangles. Dots mark vertices; one cell $c$ is shaded.}
\label{fig:2d_tessellation}
\end{figure}

Let $\Omega\subseteq\mathbb{R}^2$ be a rectangular domain partitioned into finitely many triangles. Within each triangle $c$, the velocity field takes the form
\begin{align}
v_c(p)=A_c p+b_c,
\end{align}
where $A_c\in\mathbb{R}^{2\times 2}$ and $b_c\in\mathbb{R}^2$. Continuity of $v$ is enforced across all shared cell boundaries.
For example, \autoref{fig:2d_tessellation} illustrates a tessellation of $\Omega=[0,4]\times[0,2]$ into a $4\times 2$ grid of squares, each divided into four triangles.

As in 1D, a CPA velocity field on a finite tessellation in 2D is Lipschitz continuous. Consequently, for every time at which the flow is defined, the corresponding flow map is invertible and bi-Lipschitz. In particular, it is a homeomorphism, and both the flow map and its inverse are differentiable almost everywhere.

We now explore when it is also a diffeomorphism.
The situation is more involved than in 1D because the velocity is now vector-valued, and trajectories may cross cell interfaces transversely, touch them tangentially, or move along them. Moreover, several cells may meet at a single vertex.

As in 1D, the key issue is not necessarily the discontinuity of the derivative of the velocity field, but rather how trajectories interact with the interfaces at which such discontinuities occur.

To illustrate a difficulty that does not arise in 1D, consider the CPA velocity field
\begin{align}
v(x,y)=(1,|y|).
\end{align}
This velocity field is nonzero everywhere, but its derivative is discontinuous across the interface $y=0$. The corresponding time-one flow is
\begin{align}
\Phi(x,y)=
\begin{cases}
(x+1,ey), & y\geq 0,\\
(x+1,e^{-1}y), & y<0.
\end{cases}
\end{align}
Here $\Phi=(\Phi_1,\Phi_2)$, so $\Phi_1$ and $\Phi_2$ give the final $x$- and $y$-coordinates, respectively.
For this field, $\Phi_2(x_0,y_0)$ is independent of $x_0$, so it can be plotted as a function of $y_0$ alone: the same graph applies for every initial horizontal coordinate $x_0$.

\begin{figure}[!htbp]
\centering
\includegraphics[width=\linewidth]{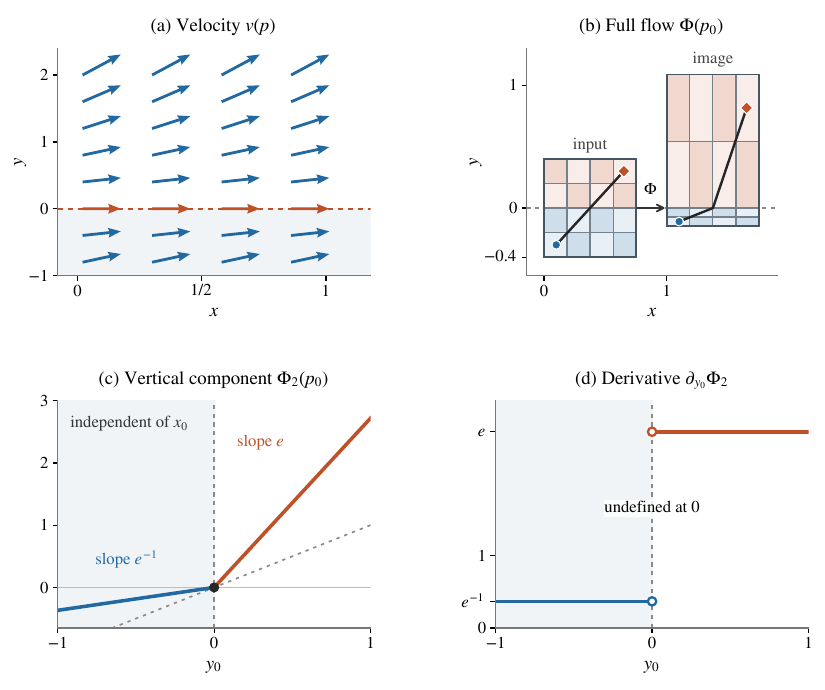}
\caption{\textbf{Loss of regularity for $v(x,y)=(1,|y|)$.}
The nonzero field is tangent to $y=0$ (a).
In (b), $\Phi$ maps the left grid and diagonal segment to the right; matching markers identify corresponding points.
The resulting flow is not differentiable at $y=0$, as illustrated by the kink created in the image of the diagonal segment.
The vertical component $\Phi_2(x_0,y_0)=y^{(1)}$ is independent of $x_0$, so (c) plots it against $y_0$ alone.
Its one-sided derivatives with respect to $y_0$ are unequal: $e^{-1}$ and $e$ (d, open circles).}
\label{fig:2d_regularity}
\end{figure}

Thus, $\Phi$ is not differentiable along $y=0$, even though the velocity field never vanishes. The issue is that the velocity is tangent to the interface, and trajectories starting on it remain there. This shows that, unlike in 1D, excluding zeros of the velocity field at cell interfaces is not sufficient.
\autoref{fig:2d_regularity} illustrates this loss of differentiability.

\paragraph{A sufficient regularity condition.}
Consider an interior edge $E$ shared by two adjacent triangles. Let $u_E$ and $w_E$ denote its endpoints, and let $n_E$ be a unit normal vector to the edge.

We impose the following two conditions on every interior edge:
\begin{align}
\boxed{
\begin{aligned}
v(p)&\neq 0,
&&\forall p\in E\cap\operatorname{int}(\Omega),\\
n_E^\top v|_E&\not\equiv 0.
\end{aligned}
}
\label{eq:2d_regularity}
\end{align}

The first condition extends the nonvanishing-velocity
requirement from the 1D case: it excludes zeros of the
velocity field on interior cell interfaces, while still
allowing zeros strictly inside cells and at the domain boundary.

The second condition addresses an additional phenomenon
that does not arise in 1D: trajectories may remain on an
interface even when the velocity is nonzero. It excludes
interfaces along which the velocity is everywhere tangent.
Importantly, it does not require the normal component of
the velocity to be nonzero everywhere. Thus, trajectories
may cross an interface transversely or touch it tangentially.

Since $v$ is affine along each edge, the second condition is equivalent to requiring that
\begin{align}
n_E^\top v(u_E)\neq 0
\quad\text{or}\quad
n_E^\top v(w_E)\neq 0.
\end{align}

These conditions are stronger than necessary. For example, an interface at which the affine velocity fields on both sides agree introduces no regularity issue, regardless of whether the conditions above hold. Nevertheless, we adopt these stronger conditions because they are simple, easy to verify, and, as shown below, satisfied almost surely under standard sampling of CPA parameters.

\paragraph{Regularity of the flow.}
We now show that the conditions above are sufficient to guarantee that the resulting CPAB transformation is a diffeomorphism.

\medskip
\noindent\textbf{Proposition.}
Suppose that $v$ is a CPA velocity field satisfying~\autoref{eq:2d_regularity} on every interior edge of the tessellation. Fix $T>0$, and let $U_T\subseteq\operatorname{int}(\Omega)$ denote the set of initial conditions whose trajectories remain in $\operatorname{int}(\Omega)$ throughout $[0,T]$. Since each such trajectory is a compact subset of $\operatorname{int}(\Omega)$ over $[0,T]$, continuous dependence on initial conditions implies that $U_T$ is open. Moreover, $\Phi^{(T)}(U_T)$ is precisely the set of points whose backward trajectories remain in $\operatorname{int}(\Omega)$ throughout $[0,T]$. The same argument therefore shows that $\Phi^{(T)}(U_T)$ is open. Then the flow map
\begin{align}
\Phi^{(T)}:U_T\longrightarrow\Phi^{(T)}(U_T)
\end{align}
is a diffeomorphism.

\medskip
\noindent\textit{Proof.}
The central observation is that, under~\autoref{eq:2d_regularity}, every trajectory that remains in the interior of the domain spends zero time on the union of interior cell interfaces.

To establish this, consider an interior edge $E$. Since $v$ is affine along $E$, its normal component
\begin{align}
g(p)=n_E^\top v(p),\qquad p\in E,
\end{align}
is an affine scalar function along the edge.

By the second condition in~\autoref{eq:2d_regularity}, $g$ is not identically zero. Therefore, it can vanish at most at one point of $E$.

Now consider a trajectory $p^{(t)}$. Whenever the trajectory crosses $E$ transversely, its intersection with the edge occurs at an isolated time.

More generally, suppose, for contradiction, that the set of times at which the trajectory lies on $E$ has positive measure.

At almost every such time, the derivative of the signed distance from the trajectory to the line containing $E$ must vanish. Consequently,
\begin{align}
n_E^\top v(p^{(t)})=0
\end{align}
at almost every such time.

Since $g$ can vanish at most at one point of $E$, the trajectory would have to occupy that same point for a set of times of positive measure.

If that point belongs to the interior of $\Omega$, the first condition in~\autoref{eq:2d_regularity} guarantees that the velocity there is nonzero. A trajectory cannot occupy a point with nonzero velocity for a set of times of positive measure. If the point belongs to the domain boundary, it cannot be reached by a trajectory that remains in the interior.

In either case, we obtain a contradiction. Thus, the trajectory spends zero time on $E$.

Since the tessellation contains finitely many edges, every interior trajectory spends zero time on the union of all interior cell interfaces. This argument also covers vertices where several triangles meet, without requiring separate treatment of the different configurations.

We now turn to differentiability of the flow.

Fix $p_0\in U_T$, and let
\begin{align}
p(t)=\Phi^{(t)}(p_0)
\end{align}
denote its trajectory. Since the trajectory spends zero time on cell interfaces, the matrix
\begin{align}
A(t)=A_{c(t)}
\end{align}
is well-defined for almost every $t\in[0,T]$, where $c(t)$ denotes the cell containing $p(t)$. We may assign arbitrary bounded values to $A(t)$ at the remaining times.

Consider the matrix-valued initial-value problem
\begin{align}
\frac{dJ(t)}{dt}
=
A(t)J(t),
\qquad J(0)=I.
\label{eq:2d_variational}
\end{align}
Since $A(t)$ is bounded and measurable, this problem has a unique absolutely continuous solution.

We will show that
\begin{align}
J(t)=D\Phi^{(t)}(p_0).
\end{align}

Consider a nearby initial condition $p_0+h$, with $h$ sufficiently small that $p_0+h\in U_T$, and denote its trajectory by
\begin{align}
p_h(t)=\Phi^{(t)}(p_0+h).
\end{align}
Define the difference between the two trajectories as
\begin{align}
\delta_h(t)=p_h(t)-p(t).
\end{align}

Since $v$ is Lipschitz continuous, with Lipschitz constant $L$, Grönwall's inequality gives
\begin{align}
\|\delta_h(t)\|
\leq e^{Lt}\|h\|,
\qquad t\in[0,T].
\label{eq:2d_flow_bound}
\end{align}

We now define the linearization error
\begin{align}
r_h(t)
=
v(p_h(t))-v(p(t))-A(t)\delta_h(t).
\label{eq:2d_remainder}
\end{align}

At almost every time $t$, the reference trajectory $p(t)$ lies strictly inside a cell. Since $p_h(t)\to p(t)$ as $h\to 0$, for sufficiently small $h$ the two points lie in the same cell. Within that cell, $v$ is affine, so
\begin{align}
r_h(t)=0
\end{align}
for all sufficiently small $h$, at almost every fixed $t$.

Moreover, let $M=\max_c\|A_c\|$. By the Lipschitz continuity of $v$ and~\autoref{eq:2d_flow_bound},
\begin{align}
\|r_h(t)\|
&\leq (L+M)\|\delta_h(t)\|\\
&\leq (L+M)e^{LT}\|h\|.
\end{align}

Thus, $\|r_h(t)\|/\|h\|$ is uniformly bounded and converges to zero for almost every $t$. By the dominated convergence theorem,
\begin{align}
\int_0^T\|r_h(t)\|\,dt
=
o(\|h\|).
\label{eq:2d_remainder_bound}
\end{align}

Using the integral form of the ODE and the definition of $r_h$, we obtain
\begin{align}
\delta_h(t)
=
h+\int_0^t A(s)\delta_h(s)\,ds
+\int_0^t r_h(s)\,ds.
\end{align}

On the other hand, the solution of~\autoref{eq:2d_variational} satisfies
\begin{align}
J(t)h
=
h+\int_0^t A(s)J(s)h\,ds.
\end{align}

Subtracting these equations and applying Grönwall's inequality yields
\begin{align}
\sup_{t\in[0,T]}
\|\delta_h(t)-J(t)h\|
&\leq
e^{MT}\int_0^T\|r_h(s)\|\,ds\\
&=o(\|h\|).
\end{align}

Consequently, the flow is differentiable with respect to its initial condition, and
\begin{align}
D\Phi^{(t)}(p_0)=J(t).
\end{align}

We have therefore established that $\Phi^{(T)}$ is differentiable at every $p_0\in U_T$. It remains to verify differentiability of the inverse.

Let $F=\Phi^{(T)}$. Since the flow is bi-Lipschitz, there exists $m>0$ such that $\|F(p)-F(q)\|\geq m\|p-q\|$ for all $p,q\in U_T$. Taking $p=p_0$ and $q=p_0+\varepsilon h$, dividing by $\varepsilon>0$, and letting $\varepsilon\to 0$ yields
\begin{align}
\|DF(p_0)h\|\geq m\|h\|,
\qquad h\in\mathbb{R}^2.
\end{align}
Thus $DF(p_0)$ is nonsingular.

Now write $y_0=F(p_0)$ and $y=F(p_0+h)$. By differentiability, $y-y_0=DF(p_0)h+o(\|h\|)$. Since the inverse is Lipschitz, $\|h\|=O(\|y-y_0\|)$, and hence
\begin{align}
F^{-1}(y)-F^{-1}(y_0)
=DF(p_0)^{-1}(y-y_0)+o(\|y-y_0\|).
\end{align}
Therefore, the inverse is differentiable as well, and $\Phi^{(T)}$ is a diffeomorphism from $U_T$ onto its image.
\hfill$\square$

\paragraph{A remark on stronger regularity.}
In fact, the preceding proof can be extended to establish
that the Jacobian of the flow is continuous with respect to
the initial condition. Thus, under the same regularity
condition, the resulting transformation is not merely a
diffeomorphism but a $C^1$ diffeomorphism.
We omit the additional details. 

\paragraph{Almost-sure satisfaction of the regularity condition.}
We now consider the tessellations commonly used in CPAB implementations~\cite{cpab,cpab2017,libcpab,Aflalo:NeurIPS:2026}.

Let $\Omega$ be a rectangle partitioned into a regular grid of squares, where each square is divided into four triangles by its two diagonals, as in \autoref{fig:2d_tessellation}. Thus, each square contains an additional vertex at its center.

The CPA velocity field is specified by its values at the mesh vertices, subject to continuity across cells. Optionally, zero boundary conditions may be imposed by fixing the velocities at the domain-boundary vertices to zero.

Assume that every interior edge has at least one endpoint whose velocity is a free parameter. This holds for the four-triangle-per-square construction on rectangular grids with at least two squares in each coordinate direction.

We claim that if the free vertex velocities are sampled jointly from an absolutely continuous distribution, then~\autoref{eq:2d_regularity} holds with probability one.

To see this, consider an interior edge $E$ with endpoints $u_E$ and $w_E$.

The second condition fails only when
\begin{align}
n_E^\top v(u_E)
=
n_E^\top v(w_E)
=
0.
\end{align}

Since at least one endpoint velocity is a free parameter, this event imposes a nontrivial linear equality on the free velocity parameters. Under an absolutely continuous distribution, it therefore has probability zero.

For the first condition, recall that the velocity along an edge is obtained by linear interpolation of its endpoint velocities.

If both endpoint velocities are free, the probability that their interpolated values vanish anywhere on the edge is zero. Indeed, such an event requires the two velocity vectors to be linearly dependent, which occurs with probability zero under an absolutely continuous joint distribution.

If one endpoint lies on the domain boundary and has its velocity fixed to zero, the interpolated velocity can vanish at that endpoint. However, it is nonzero everywhere else along the edge provided that the other endpoint velocity is nonzero, which again holds almost surely.

Thus, both conditions hold almost surely for each interior edge. Since the tessellation contains finitely many edges, they hold simultaneously for all interior edges with probability one.

In particular, the same result applies if we independently add Gaussian noise, with arbitrarily small positive variance, to the free vertex velocities while keeping the boundary velocities fixed.

Therefore, for the tessellations described above, the CPAB transformation is a diffeomorphism on the interior of the domain almost surely, provided the flow is defined there for the time interval under consideration.

\paragraph{Zero boundary conditions.}
Finally, suppose that zero boundary conditions are imposed, so that
\begin{align}
v(p)=0,
\qquad p\in\partial\Omega.
\end{align}

Every boundary point is then a fixed point of the flow. Moreover, uniqueness of ODE solutions prevents an interior trajectory from reaching the boundary in finite time. Consequently, the flow maps the interior of $\Omega$ onto itself.

In particular, the set $U_T$ in the proposition is then the entire interior of $\Omega$. Under the regularity condition above, the resulting CPAB transformation is therefore a diffeomorphism from $\operatorname{int}(\Omega)$ onto itself.

On the entire closed domain $\Omega$, the flow remains a bi-Lipschitz homeomorphism. In particular, it extends continuously to the boundary, where it equals the identity.

We do not require the derivative of the flow to extend continuously to every boundary point. Such an extension need not exist at boundary vertices where several affine pieces meet, even though the transformation is a diffeomorphism throughout the interior.

\section{The 3D Case}
\label{sec:3d}

The argument establishing differentiability of the flow in
the 2D case extends directly to 3D. Here, the domain is
tessellated into tetrahedra, and the interfaces between
adjacent cells are triangular faces. As in 2D, a sufficient
regularity condition is that every trajectory that remains in the interior of the domain spends zero time on interior interfaces
where the derivative of the CPA velocity field is
discontinuous. Under this condition, the differentiability
argument from the previous section applies without
modification. Together with the bi-Lipschitz property of
the flow, this guarantees that, for every $T>0$, the flow map $\Phi^{(T)}$ is a diffeomorphism from $U_T$ onto its image, with $U_T$ defined as in the 2D Proposition. We omit the proof.

A separate question is whether this regularity condition
is typically satisfied in practice. In 3D, the set of
points where the velocity is tangent to a cell interface
may form an entire line rather than an isolated point,
making the analysis more involved than in 2D.

For the standard five-tetrahedron subdivisions of
rectangular grids used in CPAB implementations, including
with zero-normal-velocity boundary constraints, we expect
the regularity condition to hold with probability one when
the free velocity parameters are sampled from a probability
distribution with a joint density (e.g., a Gaussian).
However, we do not establish this probability-one result
here.

\section{The mistake in~\cite{cpab,cpab2017}}

In the original CPAB papers by Freifeld et al.~\cite{cpab,cpab2017},
the pathological cases discussed in the preceding sections were
overlooked, and it was stated,
without qualification, that every CPAB transformation is a
diffeomorphism. 
The error  there, in the proof of Part (iii) of Theorem 1 from ~\cite{cpab2017},
can be traced to an incorrect statement preceding
the proof: that a Lipschitz continuous velocity field necessarily
generates a differentiable flow. This implication is false, as
demonstrated by the counterexamples in the preceding sections of the present paper.

Importantly, the invertibility argument in the original papers remains valid, although the differentiability claim should have been demoted to bi-Lipschitz continuity. As explained in the present paper, CPAB transformations are always bi-Lipschitz homeomorphisms, and hence differentiable almost everywhere, with almost-everywhere differentiable inverses. Moreover, under the sufficient regularity conditions established above, they are indeed diffeomorphisms.

Finally, these regularity conditions are not merely theoretical.
In 1D and for the 2D tessellations considered above, they can be 
easily enforced and, moreover, hold with probability one when the free
CPA parameters are sampled from any probability distribution
with a joint density (e.g., a Gaussian).
Thus, although the original unconditional claim was incorrect,
CPAB transformations are typically diffeomorphisms in these
settings. As mentioned above, 
we expect analogous behavior in 3D.

\newpage

\bibliographystyle{plainnat}
\bibliography{references}

\end{document}